\documentclass{article}

\usepackage[preprint]{neurips_2023}

\usepackage[utf8]{inputenc} % allow utf-8 input
\usepackage[T1]{fontenc}    % use 8-bit T1 fonts
\usepackage{hyperref}       % hyperlinks
\usepackage{url}            % simple URL typesetting
\usepackage{booktabs}       % professional-quality tables
\usepackage{amsfonts}       % blackboard math symbols
\usepackage{nicefrac}       % compact symbols for 1/2, etc.
\usepackage{microtype}      % microtypography
\usepackage{xcolor}         % colors
\usepackage{graphicx}
\usepackage{multirow}
\usepackage{tabularx}

\title{Understanding Counterspeech for \\Online Harm Mitigation}

\author{
Yi-Ling Chung\textsuperscript{1} \quad
Gavin Abercrombie\textsuperscript{2} \quad
Florence Enock\textsuperscript{1} \\
{
\bf Jonathan Bright\textsuperscript{1} \quad
\bf Verena Rieser\textsuperscript{2}\thanks{Now at Google DeepMind}
}\\ \\
\textsuperscript{1}The Alan Turing Institute 
\textsuperscript{2}The Interaction Lab, Heriot-Watt University \\
\{ychung, fenock, jbright\}@turing.ac.uk \\  
  \{g.abercrombie, v.t.rieser\}@hw.ac.uk 
}

\begin{document}

\maketitle

\begin{abstract}
Counterspeech offers direct rebuttals to hateful speech by challenging perpetrators of hate and showing support to targets of abuse. It provides a promising alternative to more contentious measures, such as content moderation and deplatforming, by contributing a greater amount of positive online speech rather than attempting to mitigate harmful content through removal. Advances in the development of large language models mean that the process of producing counterspeech could be made more efficient by automating its generation, which would enable large-scale online campaigns. However, we currently lack a systematic understanding of several important factors relating to the efficacy of counterspeech for hate mitigation, such as which types of counterspeech are most effective, what are the optimal conditions for implementation, and which specific effects of hate it can best ameliorate. This paper aims to fill this gap by systematically reviewing counterspeech research in the social sciences and comparing methodologies and findings with computer science efforts in automatic counterspeech generation. By taking this multi-disciplinary view, we identify promising future directions in both fields.
\end{abstract}

\section{Introduction}

The exposure of social media users to online hate and abuse continues to be a cause for public concern. Volumes of abuse on social media continue to be significant in absolute terms \citep{vidgen2019abuse}, and some claim they are rising on platforms such as Twitter where at the same time content moderation appears to be becoming less of a priority \citep{frenkel_hate_2022}. Receiving abuse can have negative effects on the mental health of targets, and also on others witnessing it \citep{siegel2020online, saha_prevalence_2019}. In the context of public figures the impact on the witnesses (bystanders) is arguably even more important, as the abuse is potentially witnessed by a large volume of people. In addition, politicians and other prominent actors are driven out of the public sphere precisely because of the vitriol they receive on a daily basis \citep{noauthor_mps_2018}, raising concerns for the overall health of democracy.  

\begin{figure}[ht!]
    \centering
    \includegraphics[width=0.75\columnwidth]{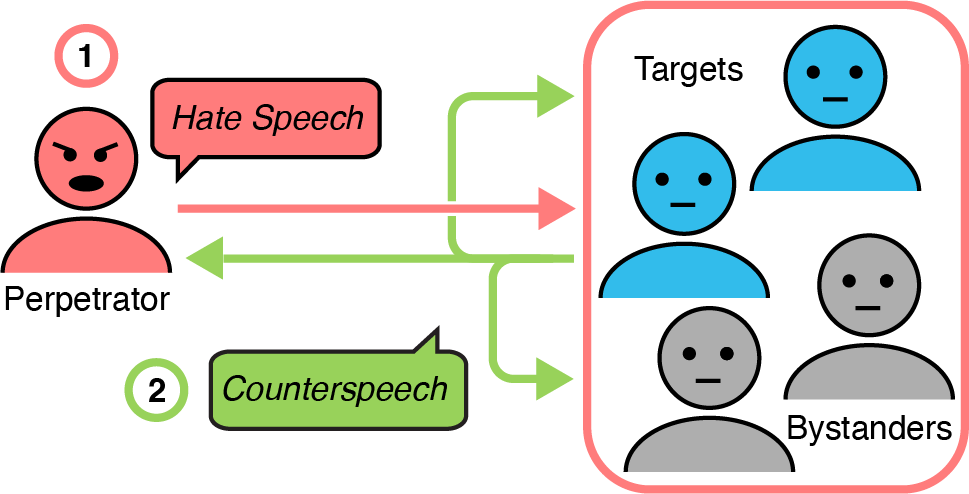}
    \caption{Counterspeech dynamics. (1) Perpetrator(s) generate Hate Speech. This may be witnessed by either targets and/or bystanders. (2) Counterspeaker(s) respond with counterspeech, which may be directed at the perpetrator(s), bystanders (e.g. to provide alternative perspectives), or other targets (e.g. in support). Counterspeakers may themselves be targets or bystanders, or could be members of organised counterspeech groups. They can have \emph{in-} or \emph{out-}group identities with respect to either the perpetrator(s) or the target(s). Counterspeech is directed at recipients, who can be one or more of (a) the perpetrator(s), (b) the target(s), or (c) other bystanders. Both counterspeakers and targets can be individual or multiple (one-to-one, one-to-many and so on).}
    \label{fig:cs-dynamics}
\end{figure}

Within this context, research on mechanisms for combating online abuse is becoming ever more important. One such research angle is the area of ``counterspeech''  (or counter-narratives): content that is designed to resist or contradict abusive or hateful content \citep{benesch2014countering, saltman2014white, bartlett2015counter},  also see Figure \ref{fig:cs-dynamics}. Such counterspeech (as we will elaborate more fully below) is an important potential tool in the fight against online hate and abuse as it does not require any interventions from the platform or from law enforcement, and may contribute to mitigating the effects of abuse \citep{munger2017tweetment, doi:10.1177/20563051211063843, doi:10.1073/pnas.2116310118, doi.org/10.1002/ab.21948} without impinging on free speech. Several civil organisations have used counterspeech to directly challenge hate, and Facebook has launched campaigns with local communities and policymakers to promote accessibility to counterspeech tools.\footnote{\url{https://counterspeech.fb.com/en/}} Similarly, Moonshot and Jigsaw implemented The Redirect Method, presenting alternative counterspeech or counter videos when users search queries that may suggest an inclination towards extremist content or groups.\footnote{\url{https://moonshotteam.com/the-redirect-method/}} 

The detection and generation of counterspeech is important because it underpins the promise of AI-powered assistive tools for hate mitigation. Identifying counterspeech is vital also for analytical research in the area: for instance, to disentangle the dynamics of perpetrators, victims and bystanders \citep{mathew2018analyzing, garland-etal-2020-countering, garland2022impact}, as well as determining which responses are most effective in combating hate speech \citep{mathew2018analyzing, mathew2019thou, chung-etal-2021-multilingual}. 

Automatically producing counterspeech is a timely and important task for two reasons. First, composing counterspeech is time-consuming and requires considerable expertise to be effective \citep{CHUNG2021100150}. Recently, large language models have been able to produce fluent and personalised arguments tailored to user expectations addressing various topics and tasks. Thus, developing counterspeech tools is feasible and can provide support to civil organisations, practitioners and stakeholders in hate intervention at scale. Second, by partially automating counterspeech writing, such assistive tools can lessen practitioners' psychological strain resulting from prolonged exposure to harmful content \citep{RIEDL2020106262, CHUNG2021100150}.

However, despite the potential for counterspeech, and the growing body of work in this area, the research agenda remains a relatively new one, which also suffers from the fact that it is divided into a number of disciplinary silos. In methodological terms, meanwhile, social scientists studying the dynamics and impacts of counterspeech \cite[e.g.][]{munger2017tweetment, doi:10.1177/20563051211063843, doi:10.1073/pnas.2116310118, doi.org/10.1002/ab.21948} often do not engage with computer scientists developing models to detect and generate such speech \cite[e.g.][]{chung-etal-2021-towards, Saha2022} (or vice versa). 

The aim of this review article is to fill this gap, by providing a comprehensive, multi-disciplinary overview of the field of counterspeech covering computer science and the social sciences over the last ten years. We make a number of contributions in particular. Firstly, we outline a definition of counterspeech and a framework for understanding its use and impact, as well as a detailed taxonomy. We review research on the effectiveness of counterspeech, bringing together perspectives on the impact it makes when it is experienced. We also analyse technical work on counterspeech, looking specifically at the task of counterspeech generation, scalability, and the availability and methodology behind different datasets. Importantly, across all studies, we focus on commonalities and differences between computer science and the social sciences, including how the impact of counterspeech is evaluated and which specific effect of hate speech it best ameliorates.

We draw on our findings to discuss the challenges and directions of open science (and safe AI) for online hate mitigation. We provide evidence-based recommendations for automatic approaches to counterspeech tools using Natural Language Processing (NLP). Similarly, for social scientists, we set out future perspectives on interdisciplinary collaborations with AI researchers on mitigating online harms, including conducting large-scale analyses and evaluating the impact of automated interventions. Taken together, our work offers researchers, policy-makers and practitioners the tools to further understand the potentials of automated counterspeech for online hate mitigation.

\section{Background}
Interest in investigating the social and computational aspects of counterspeech has grown considerably in the past five years. However, while extant work reviews the impact of counterspeech on hate mitigation \citep{saltman2014white, carthy2020counter, buerger2021counterspeech}, none have systematically addressed this issue in combination with computational studies in order to synthesise social scientific insights and discuss the potential role of automated methods in reducing harms. \citet{carthy2020counter} present a focused (2016-2018) systematic review of research into the impact of counter-narratives on prevention of violent radicalisation. 
They categorise the techniques employed in counter-narratives into four groups: (1) counter-stereotypical exemplars (challenging stereotypes, social schema or moral exemplars), (2) persuasion (e.g., through role-playing and emotion inducement), (3) inoculation (proactively reinforcing resistance to attitude change or persuasion), and (4) alternative accounts (disrupting false beliefs by offering different perspectives of events). The measurements of counter-narrative interventions are based on (1) intent of violent behaviour, (2) perceived symbolic/realistic group threat (e.g., perception of an out-group as dangerous), and (3) in‐group favouritism/out‐group hostility (e.g., level of trust, confidence, discomfort and forgiveness towards out-groups). They argue that counter-narratives show promise in reducing violent radicalisation, while its effects vary across techniques, with counter-stereotypical exemplars, inoculation and alternative accounts demonstrating the most noticeable outcomes. 
\citet{buerger2021counterspeech} reviews the research into the effectiveness of counterspeech, attempting to categorise different forms of counterspeech, summarise the source of influences in abusive/positive behaviour change, and elucidate the reasons which drive strangers to intervene in cyberbullying. Here, the impact of counterspeech is mostly evaluated by the people involved in hateful discussions, including hateful speakers, audiences, and counterspeakers. In comparison, we focus on \textit{what} makes counterspeech effective by comprehensively examining its use based on aspects such as strategies, audience and evaluation.  

On the computational side, some work reviews the use of counterspeech in social media using natural language processing, including work outlining counterspeech datasets \citep{adak2022mining, alsagheer2022counter}, discussing automated approaches to counterspeech classification \citep{alsagheer2022counter} and generation \citep{chaudhary2021countering, alsagheer2022counter}, and work focusing on system evaluation \citep{alsagheer2022counter}. However, work from computer sciences is not typically informed by important insights from the social sciences, including the key roles of intergroup dynamics, the social context in which counterspeech is employed, and the mode of persuasion by which counterspeech operates. Taking an interdisciplinary approach, we join work from the computer and social sciences.

\section{Review Methodology}

\begin{figure}[ht!]
    \centering
    \includegraphics{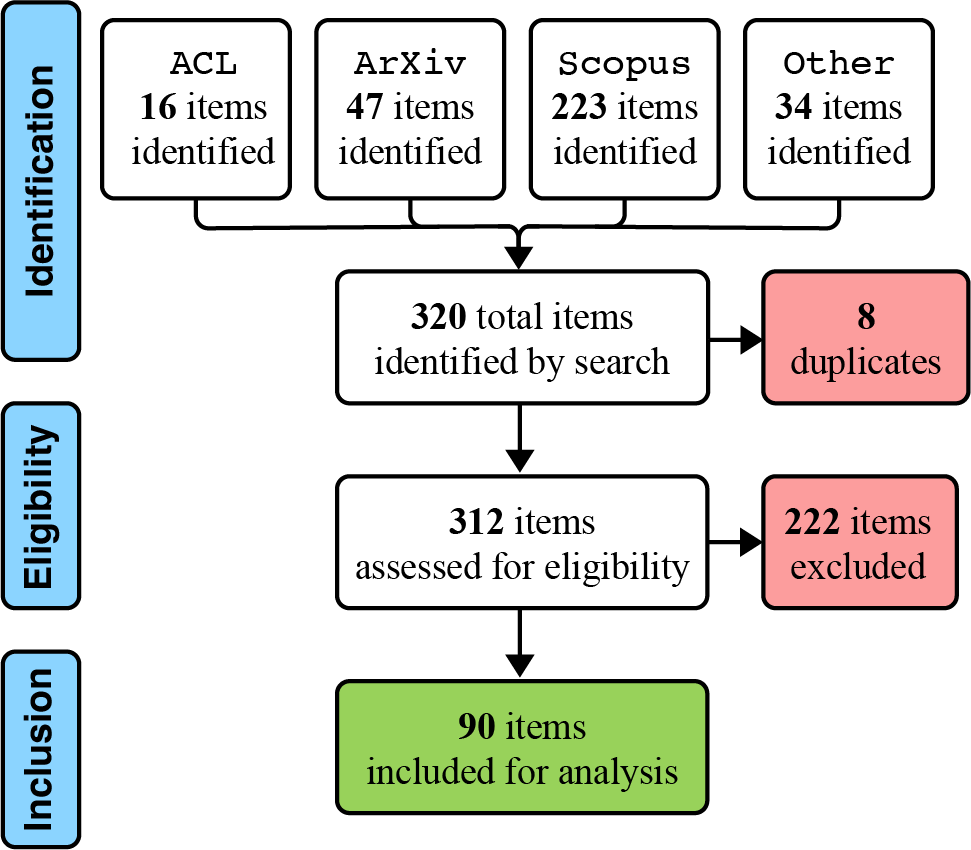}
    \caption{Flow diagram showing the identification, eligibility screening, and inclusion 
    phases of the selection of items analysed in this review.}
    \label{fig:prisma}
\end{figure}

Taking a multi-disciplinary perspective, we systematically review work on counterspeech from computer science and the social sciences published in the past ten years. To ensure broad coverage and to conduct a reproducible review, we follow the systematic methodology of \citet{moher-etal-2009-preferred}. The search and inclusion process is shown in Figure \ref{fig:prisma}.

We used keyword terms related to counterspeech to search three key databases (ACL Anthology, ArXiv, and Scopus) that together offer a broad coverage of our target literature. We included the search terms `counter-speech', `counter-narratives', `counter-terrorism', `counter-aggression', `counter-hate', `counter speech', `counter narrative', `countering online hate speech', `counter hate speech', and `counter-hate speech'.
We also included 34 publications that we had identified previously from other sources, but that were not returned by keyword search due to not including relevant keywords or not being indexed in the target search repositories.
The search was conducted in December 2022.
Of the returned results, we include all publications that concern (1) analysis of the use and effectiveness of interventions against hateful or abusive language online, (2) characteristics of counterspeech users or recipients, or (3) data and/or implementation designed for counterspeech (e.g., counterspeech classification or generation).
These inclusion criteria were applied by two of the authors.
Following this process, we include 90 papers for analysis in this review. Each of the papers was read by at least one of the co-authors of the article.

Our review is divided into several sections (the results of which are presented sequentially below). First, we examine definitional characteristics of counterspeech, looking at how the term itself is defined, how different taxonomies have been created to classify different types of counterspeech, and the different potential purposes attributed to it. Then, we examine studies that have looked at the impact of counterspeech, discussing the different analytical designs employed and analysing evidence of the results. Following this, we discuss computational approaches to counterspeech, focusing in particular on both detection and generation. Finally, we examine ethical issues in the domain of counterspeech, and also speculate about future perspectives and directions in the field. 

\section{Defining counterspeech}
Counterspeech is multifaceted and can be characterised in several different ways. In Table \ref{table:cs_framework} we outline a framework for describing and designing counterspeech, covering who (speaker) sends what kinds of messages (strategies) to whom (recipients), and for what purpose (purpose).  
Using this structure, we summarise how counterspeech has typically been categorised in past studies. 

\begin{table*}[ht!]
\centering
\begin{tabularx}{\textwidth}{l|p{11.5cm}}
\toprule
Aspects         &   Description  \\
\hline
Speaker & Who is the counterspeaker? What is the social identity and status of the counterspeaker? \\
Strategy & Which linguistic and rhetorical methods are used in the counterspeech? Which emotions or attitudes are expressed towards the hateful content? \\
Recipient & Who is the target audience? Are they hate speakers, targets of hate, or bystanders? \\
Purpose & What is the aim of disseminating counterspeech?\\
\bottomrule
\end{tabularx}
\caption{Framework for describing and designing counterspeech.}
  \label{table:cs_framework}
\end{table*}

Most studies in the field use one of three main terms: \textit{counterspeech}, \textit{counter-narratives} \citep{reynolds2016counter, doi:10.1080/09546553.2021.1962308, tuck2016counter, doi:10.1080/18335330.2019.1574020} and \textit{hope speech} \citep{10.1093/oxfordhb/9780199399314.013.3}. These three terms broadly refer to a similar concept: content that challenges and rebuts hateful discourse and propaganda \citep{saltman2014white, bartlett2015counter, benesch2016, doi:10.1080/1057610X.2021.1888404, garland2022impact} using non-aggressive speech \citep{benesch2016, reynolds2016counter, schieb2016governing}. There are some differences between the terms. \citet{ferguson2016countering} considers counter-narratives as intentional strategic communication within a political, policy, or military context. Additionally, the term counter-narrative also refers to narratives that challenge a much broader view or category such as forms of education,  propaganda, and public information \citep{benesch2016}. Such counter-narratives are often discussed in the context of the prevention of violent extremism. Hope speech, meanwhile, could be seen as a particular type of counterspeech: it promotes positive engagement in online discourse to lessen the consequences of abuse, and places a particular emphasis on delivering optimism, resilience, and the values of equality, diversity and inclusion \citep{PMID:35844297}. In this paper, we review work that relates to all of these three concepts, and largely make use of the catch-all term counterspeech, while acknowledging the slight differences between the concepts. 

\subsection{Classifying counterspeech} \label{sec:existing_taxonomy}

Researchers have identified a variety of different types of counterspeech. Here, we outline four main ways in which counterspeech can vary, in terms of the identity of the counterspeaker, the strategies employed, the recipient of the counterspeech and the purpose of counterspeech. 

\paragraph{Counterspeakers (who)}
Psychological studies show that the identity of a speaker plays a key role in how large an audience their message reaches and how persuasive the message is. Common crucial factors include group identity (such as race, religion, and nationality), level of influence, and socioeconomic status. For instance, counterspeech provided by users with large numbers of followers and from an in-group member is more likely to lead to changes in the behaviour of perpetrators of hate~\citep{munger2017tweetment}. 

Some studies characterise individuals who use counterspeech and suggest that these users exhibit different characteristics and interests than users who spread hate \citep{mathew2018analyzing, mathew2019thou, doi:10.1177/20563051211063843}. Through lexical, linguistic and psycholinguistic analysis of users who generate hate speech or counterspeech on Twitter, \citet{mathew2018analyzing} find that counterspeakers are higher in agreeableness, displaying traits such as altruism, modesty, and sympathy, and display higher levels of self-discipline and conscientiousness. Possibly driven by a motive to help combat hate speech, counterspeakers tend to use words related to government, law, leadership, pride, and religion. Regarding the impact of being a counterspeaker, in an ethnographic study, members of a counterspeech campaign reported feeling more courageous and keen to engage in challenging discussions after expressing opinions publicly \citep{doi:10.1177/20563051211063843}.

\paragraph{Strategies (how)} Counterspeech can take many forms. 
\citet{benesch2016} first identify eight types of counterspeech used on Twitter: (1) \emph{presentation of facts}, (2) \emph{pointing out hypocrisy or contradiction}, (3) \emph{warning of consequences}, (4) \emph{affiliation} [i.e. establishing an emotional bond with the perpetrators or targets of hate], (5) \emph{denouncing}, (6) \emph{humour/sarcasm}, (7) \emph{tone} [a tendency or style adopted for communication, e.g., empathetic and hostile], and (8) \emph{use of media}.
Based on this taxonomy, follow-up studies on counterspeech make minor modifications to cover strategies in a broader scope.
\citet{mathew2018analyzing} analyzed and classified counterspeech on Twitter, taking \citet{benesch2016}'s taxonomy but dropping \emph{the use of media} and adding \textit{hostile language} and \textit{positive tone}, which replaces general strategy \emph{tone}. 
Similarly, \citet{mathew2019thou} collected and annotated counterspeech comments from Youtube, adopting \citet{benesch2016}'s taxonomy but excluding \textit{tone} and adding \textit{positive tone}, \textit{hostile language} and \textit{miscellaneous}. \citet{chung-etal-2019-conan} collaborated with NGOs to collect manually written counterspeech. For data annotation, they followed the taxonomies provided by \citet{benesch2016} and \citet{mathew2019thou}, while adding \textit{counter question} and discarding \textit{the use of media}. Counterspeech examples for each strategy are provided in Table \ref{table:cs_examples}.  

\begin{table*}[ht!]
\centering
\begin{tabularx}{\textwidth}{l|p{10.5cm}}
\toprule
Strategy         &   Example  \\
\hline
Facts & Actually, studies show that on the whole migrants contribute more to public finances than they take out, see this article for example.\\
Hypocrisy & Immigrants stealing British resources? A bit rich given how much was stolen from colonies by the British Empire.\\
Consequences  & Spreading hateful content is illegal. Police will knock on your door.\\
Affiliation  & As a British national, I know life is hard here right now. But I assure you that your unemployment is not the fault of immigrants.\\
Denouncing  & Stop with the racist and derogatory slurs. It's unacceptable to talk this way.\\
Counter questions  & Do you have a problem with all immigrants or only ones from lower income countries? Are you suggesting we have enough qualified and willing British born workers to fill all the jobs? \\
Humour  & You should think about how the Spanish feel next time you go on holiday to Costa Del Sol (laughing emoji)?\\
Positive tone  & Immigrants strengthen UK society in so many ways - greater diversity, skillsets and innovation to name a few! And no way our NHS could function without the immigrant workforce.\\
\bottomrule
\end{tabularx}
\caption{Synthetic examples of different counterspeech strategies in response to an example of abuse against immigrants. Here the abuse example is: `Immigrants are invading and stealing our resources'.}
  \label{table:cs_examples}
\end{table*}

\paragraph{Counterspeech recipients (whom)} Depending on the purpose of the counterspeech, the target audience may be perpetrators, victims or bystanders (see Figure \ref{fig:cs-dynamics}). Identifying the appropriate target audience or ‘Movable Middle’ is crucial to maximise the efficacy of counterspeech. Movable middle refers to individuals who do not yet hold firm opinions on a topic and can hence be potentially open to persuasion. They are also receptive to arguments and more willing to listen. These individuals often serve as ideal recipients of messages addressing social issues such as vaccination hesitancy \citep{vaccines10101739}. In the context of counterspeech, previous studies show that a small group of counterspeakers can shape online discussion when the audience holds moderate views \citep{schieb2016governing, doi:10.1177/20563051211063843}. 

\citet{wright2017vectors} group counterspeech acts into four categories based on the number of people involved in the discussion: \emph{one-to-one}, \emph{one-to-many}, \emph{many-to-one}, or \emph{many-to-many}. Some successful cases where counterspeech induces favourable changes in the discourse happen in a one-to-one discussion. This allows for dedicated opinion exchange over an ideology, which in some cases even yields long-lasting changes in beliefs. The use of hashtags is a good example of one-to-many and many-to-many interaction where conversations surge quickly \citep{benesch2016, wright2017vectors}. For instance, Twitter users often include hashtags to express support (e.g., \#BlackLivesMatter) or disagreement with haters (e.g., \#StopHate) to demonstrate their perspective.

\paragraph{The purpose of counterspeech}
Hateful language online can serve to reinforce prejudice \citep{citron2011intermediaries}, encourage further division, promote power of the ingroup, sway political votes, provoke or justify offline violence, and psychologically damage targets of hate \citep{jay2009offensive}. Just as the effects of hate are wide-ranging, counterspeech may be used to fulfil a variety of purposes. 

\vspace{0.2cm}

\noindent\textbf{$\bullet$ Changing the attitudes and behaviours of perpetrators}
In directly challenging hateful language, one key aim of counterspeech can be to change the attitudes of the perpetrators of hate themselves. The strategy here is often to persuade the perpetrator that their attitudes are mistaken or unacceptable, and to deconstruct, discredit or delegitimise extremist narratives and propaganda \citep{reynolds2016counter}. Counterspeech aimed at changing the attitudes of spreaders of hate may address the hate speaker directly, countering claims with facts or by employing empathy and affiliation. Challenging attitudes is often seen as a stepping stone to altering behaviours \citep{stroebe2008strategies}. In attempting to change the minds of perpetrators, counterspeakers ultimately hope to discourage associated behaviours such as sharing such content again in the future or showing support for other hateful content (i.e., stopping the spread of hate). In changing the minds of perpetrators, counterspeakers may also hope to prevent them from engaging in more extreme behaviours such as offline violence. 

\vspace{0.2cm}

\noindent\textbf{$\bullet$ Changing the attitudes and behaviours of bystanders}
More commonly, counterspeech is initiated with the intention of reaching the wider audience of bystanders rather than perpetrators of hate themselves \citep{buerger2022they}. These bystanders are not (at least yet) generating hateful language themselves, but rather are people exposed to hateful content either incidentally or by active engagement. Here, counterspeakers hope to persuade bystanders that the hateful content is wrong or unacceptable, again by deconstructing and delegitimising the hateful narrative. The strategy here may be to offer facts, point out hypocrisy, denounce the content, or use humour to discredit the speaker. Additionally, counterspeakers will often invoke empathy for targets of hate. In preventing bystanders from forming attitudes and opinions in line with the hateful narrative, counterspeakers hope to mitigate further intergroup division and related behaviours such as support for or engagement with additional abuse or physical violence. Counterspeakers may also hope to encourage others to generate rebutals and rally support for victims \citep{benesch2014countering}, bringing positive changes in online discourse. 

\vspace{0.2cm}

\noindent\textbf{$\bullet$ Showing support for targets of hate}
A third key way in which counterspeech functions is to show support directly to targets of hate. Online abuse can psychologically damage the wellbeing of targets and leave them feeling fearful, threatened, and even in doubt of their physical safety \citep{benesch2014defining, leader2016dangerous, saha_prevalence_2019, siegel2020online}. By challenging such abuse, counterspeakers can offer support to targets and encourage bystanders to do the same \citep{doi:10.1177/20563051211063843}. This support aims to alleviate negative emotion brought on by hate by demonstrating to targets that they are not alone and that many people do not hold the attitudes of the perpetrator. Here the particular strategies may be to denounce the hate and express positive sentiment towards the target group. Intergroup solidarity may in turn reduce retaliated antagonism.

\section{The Impact of Counterspeech} \label{sec:cs_impact}   
The concrete effects of using counterspeech remain debated. The methods applied for evaluating the effectiveness of counterspeech vary considerably across studies in the field. In this section, we outline eight aspects that can help to better understand the impact of counterspeech.

\paragraph{Research design}
A wide range of methodologies have been adopted to assess the impact of counterspeech on hate mitigation, including observational studies \citep{ernst2017hate, Stroud2018, garland2022impact}, experimental \citep{munger2017tweetment, doi:10.1177/14614448211017527, doi:10.1073/pnas.2116310118} and quasi-experimental designs \citep{doi.org/10.1002/ab.21948}. In observational studies, investigators typically assess the relationship between exposure to counterspeech and outcome variables of interest without any experimental manipulation. For instance, a longitudinal study of German political conversations on Twitter examined the interplay between organized hate and counterspeech groups \citep{garland2022impact}. There is also an ethnographic study interviewing counterspeakers on Facebook to understand external and internal practices for collectively intervening in hateful comments, such as how to build effective counterspeech action and keep counterspeakers engaged \citep{doi:10.1177/20563051211063843}. For experimental and quasi-experimental designs, both aim at estimating the causal effects of exposure to different kinds of counterspeech on outcome variables in comparison with controls (no exposure to counterspeech).

\paragraph{Languages and countries}
In the reviewed work, the impact of counterspeech is investigated in five different languages across nine countries. Notably, experiments are focused on counterspeech used in Indo-European languages such as English (USA, UK, Canada and Ireland), German (Germany), Urdu (Pakistan) and Swedish (Sweden). Only two studies are dedicated to Afro-Asiatic languages, Arabic (Egypt and Iraq). 
We did not find research dedicated to other language families, suggesting that the language coverage of counterspeech studies is still low. 

\paragraph{Platforms}
Most experiments were conducted on text-based social media platforms, such as eight on Twitter \citep{benesch2016, reynolds2016counter, silverman2016impact, Stroud2018, munger2017tweetment, doi:10.1073/pnas.2116310118, doi:10.1177/1461444820903319, garland2022impact}, six on Facebook \citep{reynolds2016counter, silverman2016impact, schieb2016governing, 10.5771/2192-4007-2018-4-555, doi:10.1080/1057610X.2021.1888404, doi:10.1177/20563051211063843}, and one on Reddit \citep{doi.org/10.1002/ab.21948}, as well as image-based online spaces, such as three on Youtube \citep{reynolds2016counter, silverman2016impact, ernst2017hate} and one on Instagram \citep{Stroud2018}. Often, the counterspeech interventions are directly monitored on such platforms, but in some cases, fictitious platforms are created in order to mimic online social activity under a controlled environment  \citep{doi:10.1177/14614448211017527, doi:10.1080/09546553.2021.1962308, 10.3389/fpsyg.2020.01059}. There are three studies analysing the impact of counterspeech across multiple platforms \citep{reynolds2016counter, silverman2016impact, Stroud2018}.

Twitter and Facebook are widely used for measuring the effects of counterspeech, with eight and six experiments respectively. For Twitter, this can be explained by its easily accessible API (even if at the time of writing continued research access to the API was in doubt). Similarly, because of difficulties in gathering data, \citet{schieb2016governing} resort to developing an agent-based computational model for simulating hate mitigation with counterspeech on Facebook. It is worth highlighting that none of the studies we reviewed had investigated recently popular mainstream platforms, such as Tiktok, Weibo, Telegram, and Discord.

\paragraph{The target of hate speech}
Abusive speech can be addressed towards many different potential targets, and each individual hate phenomenon may require different response strategies for maximum effectiveness. 
Existing studies have evaluated the effectiveness of counterspeech on several hate phenomena, with Islamophobia, Islamic extremism, and racism being the most commonly addressed, while hate against LGBTQ+ community and immigrants being the least studied. 
In these studies, abusive content is typically identified based on two strategies - hateful keyword matches \citep{doi:10.1073/pnas.2116310118, doi.org/10.1002/ab.21948}, or user accounts (e.g., content produced by known hate speakers) \citep{garland2022impact}.

\paragraph{Types of interventions} A wide range of methods are exploited to design and surface counterspeech messages to a target audience. We broadly categorise these methods based on modality and approach to creation. Counter speech is generally conveyed in text \citep{10.3389/fpsyg.2020.01059, doi:10.1073/pnas.2116310118, doi:10.1177/1461444820903319} or video mode \citep{ernst2017hate, doi:10.1080/1057610X.2021.1888404, doi:10.1080/09546553.2021.1962308}. In both cases, counterspeech materials can be created in three different ways: written by experimenters as stimuli \citep{doi:10.1177/14614448211017527, doi:10.1080/09546553.2021.1962308}, as well as written by individuals or campaigns that are collected from social media platforms \citep{benesch2016, garland2022impact, doi:10.1177/20563051211063843}. 
We also found one study integrating counterspeech messages in media such as films, TV dramas and movies \citep{doi:10.1080/18335330.2019.1574020}.

\paragraph{Counterspeech strategies} Following the strategies summarised in Section \ref{sec:existing_taxonomy}, commonly used counterspeech strategies include facts \citep{doi:10.1177/20563051211063843, doi:10.1177/14614448211017527}, denouncing \citep{Stroud2018, doi:10.1080/1057610X.2021.1888404}, counter-questions \citep{silverman2016impact, reynolds2016counter, doi:10.1080/1057610X.2021.1888404}, and a specific tone (humour or empathy) \citep{reynolds2016counter, munger2017tweetment, doi:10.1073/pnas.2116310118, doi:10.1080/1057610X.2021.1888404}. There are more fine-grained tactics for designing counterspeech in social science experiments. 
According to psychological studies, the use of social norms can reduce aggression and is closely related to legal regulation in society \citep{doi.org/10.1002/ab.21948}. This tactic was tested in an intervention study where participants were exposed to counterspeech with one of the inducements of empathy, descriptive norms (e.g., \textit{Let's try to express our points without hurtful language}) and prescriptive norms (e.g., \textit{Hey, this discussion could be more enjoyable for all if we would treat each other with respect.}) \citep{doi.org/10.1002/ab.21948}. \citet{10.3389/fpsyg.2020.01059} designed counterspeech based on substances rather than tactics, varying three different narratives: (1) social (seeking to establish a better society), (2) political (bringing a new world order through a global caliphate), and (3) religious (legitimising violence based on religious purposes). Considering broader counterspeech components, a few organisations further focus on challenging ideology (e.g., far-right and Islamist extremist recruitment narratives), rather than deradicalising individuals \citep{silverman2016impact, doi:10.1080/1057610X.2021.1888404}. Counterspeech drawing from personal stories in a reflective or sentimental tone is also considered as it can resonate better with target audiences \citep{silverman2016impact}. In addition to neutral or positive counterspeech, radical approaches are taken by counter-objecting, degrading or shaming perpetrators in public for unsolicited harmful content \citep{Stroud2018, doi:10.1177/14614448211017527}.

\paragraph{Types of evaluation metrics} Based on~\citet{reynolds2016counter}'s counterspeech \emph{Handbook}, we identified the following three types of metrics used by the authors of the papers to evaluate the effectiveness of counterspeech interventions: social impact, behavioural change, and attitude change measures.
\vspace{0.2cm}

\noindent\textbf{$\bullet$ Social impact metrics} are (usually automated) measurements of how subjects interact with counterspeech online. 
Such measures include, bounce rate, exit rate,\footnote{Bounce rate is the number of users who leave a website without clicking past the landing page; exit rate measures how many people leave the site from a given section~\citep{reynolds2016counter}.} geo-location analysis and the numbers of likes, views, and shares that posts receive~\citep{garland-etal-2020-countering,doi:10.1073/pnas.2116310118,doi:10.1177/1461444820903319,reynolds2016counter,10.5771/2192-4007-2018-4-555,doi:10.1080/1057610X.2021.1888404,silverman2016impact}.
For example, for one of their experiments, \citet{doi:10.1080/1057610X.2021.1888404} measure the `click-through rates' of Facebook users redirected from hateful to counterspeech materials, while \citet{doi:10.1073/pnas.2116310118} measure retweets and deletions (in addition to behavioural change measures). 

Social impact measures are also applied to synthetic data by \citet{schieb2016governing}, who measure the `likes' of their (simulated) participants as hate and counterspeech propagate through a network (as well as applying behavioural metrics). Taking a more distant, long-term view, \citet{doi:10.1080/18335330.2019.1574020} cite Egypt's overall success at countering radicalisation with counterspeech campaigns by comparing its position on the Global Terrorism Index with that of Pakistan.

While the majority of these measurements are automated, \citet{10.5771/2192-4007-2018-4-555} use survey questions to examine participants willingness to intervene against hate speech depending on the severity of  the hate, the number of bystanders, and the reactions of others.
Unlike the survey-based approaches described below, they do not consider \emph{changes} in attitude. In addition, \citet{doi:10.1177/20563051211063843} assess the success of the \#jagärhär counterspeech campaign (\#iamhere in English, a Sweden-based collective effort that has been applied in more than 16 countries) based on the extent to which it has facilitated the emergence of alternative perspectives.
\vspace{0.2cm}

\noindent\textbf{$\bullet$ Behavioural change measures} reveal whether subjects change their observable behaviour towards victims before and after exposure to counterspeech, for example in the tone of their language as measured with sentiment analysis.

For instance, \citet{doi:10.1073/pnas.2116310118} conduct sentiment analysis to determine the behaviour of previously xenophobic accounts after treatment with counterspeech, \citet{doi.org/10.1002/ab.21948} measure levels of verbal aggression before and after interventions, and \citet{garland-etal-2020-countering} assess the proportion of hate speech in online discourse before and after the intervention of an organised counterspeech group.
Other such measures are those of \citet{doi:10.1080/1057610X.2021.1888404}, who compare the number of times users violate Facebook policies before and after exposure to counterspeech, and
\citet{munger2017tweetment}, who examine the likelihood of Twitter users continuing to use racial slurs following sanctions by counterspeakers of varying status and demographics.
And in a network simulation experiment, \citet{schieb2016governing} measure the effect of positive or negative (synthetic) posts on (synthetic) user behaviour. 
\vspace{0.2cm}

\noindent\textbf{$\bullet$ Attitude change measures} are used to assess whether people (hate/counter speakers or bystanders) change their underlying attitudes or intentions through non-automated methods such as interviews, surveys, focus groups, or qualitative content analysis.

For potential hate speech perpetrators, \citet{doi:10.1080/09546553.2021.1962308} use psychological testing to measure the extent to which participants legitimized violence after exposure to differing counterspeech strategies,  \citet{10.3389/fpsyg.2020.01059} compare support for ISIS and other factors using in participants exposed to differing counterspeech strategies and a control group,
\citet{ernst2017hate} code user comments on hate and counterspeech videos to perform qualitative content analysis of users' attitudes.

For bystanders that may be potential counterspeakers, \citet{doi:10.1177/14614448211017527} use a survey to examine whether counterspeech leads to increased intentions to intervene.
And for those already engaged in counterspeech, \citet{doi:10.1177/20563051211063843} conduct interviews with members of an organised group to reveal their perceptions of the efficacy of their interventions.

\paragraph{Effectiveness}

Owing to the variation in experimental setups, aims, and evaluation methods of the counterspeech efforts we review, it is not straightforward to compare their levels of success.
Indeed, several of the studies concern broad long-term goals that cannot be easily evaluated at all~\citep[e.g.][]{reynolds2016counter,silverman2016impact} or provide only anecdotal evidence~\citep[e.g.][]{benesch2016,Stroud2018,doi:10.1177/20563051211063843}.

Beyond this, evidence of successful counterspeech forms a complex picture.
For example, \citet{garland2022impact} show that organised counterspeech is effective, but can produce backfire effects and actually attract more hate speech in some circumstances.
They also show that these dynamics can alter surrounding societal events---although they do not make causal claims for this.
Similarly, \citet{ernst2017hate} find mixed results, with counterspeech encouraging discussion about hate phenomena and targets in some cases, but also leading to increases in hateful comments.
However, \citet{silverman2016impact} suggest that even such confrontational exchanges can be viewed as positive signs of engagement.

There is some evidence for the comparative efficacy of different counterspeech strategies.
\citet{doi.org/10.1002/ab.21948} find that three of their intervention types (`disapproval', `abstract norm', `empathy') are effective in reducing verbal violence when compared with no intervention at all. 
Here, empathy had the weakest effect, which they put down to the empathetic messages being specific to particular behaviours, limiting their capacity to modify aggression towards wider targets.
\citet{doi:10.1073/pnas.2116310118} also found that empathy-based counterspeech can consistently reduce hate speech, although this effect is small.
And \citet{doi:10.1080/09546553.2021.1962308} found that counterspeech that seeks to correct false information in the hate speech actually leads to higher levels of violence legitimisation, while having participants actively counter terrorist rhetoric themselves (`Tailored Counter-Narrative') was the most effective strategy to reduce this.
They found counterspeech to be more effective on participants that are already predisposed to cognitive reflection.
However, focusing on the effect of factual correction on the victims rather than perpetrators of hate speech, \citet{doi:10.1177/14614448211017527} found it to be effective in providing support and preventing them from hating back and therefore widening the gap between groups.

There is also some evidence that the numbers of the different actors involved in a counterspeech exchange can affect an intervention's success.
\citet{schieb2016governing} find that counterspeech can impact the online behaviour of (simulated) bystanders, with the effectiveness strongly influenced by the proportions of hate and counter speakers and neutral bystanders.
According to their model, a small number of counterspeakers can be effective against smaller numbers of hate speakers in the presence of larger numbers of people lacking strong opinions.
\citet{doi:10.1080/1057610X.2021.1888404} found their counterspeech strategies to be  effective only for higher risk individuals within the target populations, although they did not see any of the potential negative effects of counterspeech (such as increased radicalisation) reported elsewhere.

Focusing on who in particular delivers counterspeech, \citet{munger2017tweetment} finds that success of counterspeech depends on the identity and status of the speaker.
However, with only a small positive effect, \citet{10.3389/fpsyg.2020.01059} found that the content of counterspeech was more important than the source.
And \citet{garland2022impact} found that, while organised counterspeech can be effective, the efforts of individuals can lead to increases in hate speech.
In \citet{doi:10.1177/20563051211063843}, members of \#jag{\"a}rh{\"a}r claim that their counterspeech interventions were successful in making space for alternative viewpoints to hate speech.

\section{Computational Approaches to Counterspeech}

In this section, we switch the focus to look at literature on counterspeech emerging from the field of computer science. We tackle three subjects in particular: the datasets being used in these studies, approaches to counterspeech detection, and approaches to counterspeech generation.  
 
\subsection{Counterspeech Datasets}
Approaches for counterspeech collection focus on gathering two different kinds of datasets: spontaneously produced comments crawled from social media platforms, and deliberately created responses aiming to contrast hate speech. In the first case, content is retrieved based on keywords/hashtags related to targets of interest \citep{mathew2018analyzing, vidgen-etal-2020-detecting, 10.1145/3487351.3488324, vidgen-etal-2021-introducing} or from pre-defined counterspeech accounts \citep{garland-etal-2020-countering}. In principle, due to the easily accessible API required for data retrieval, the majority of datasets are collected from social media platforms including Twitter \citep{mathew2018analyzing, procter2019study, garland-etal-2020-countering, kennedy2020constructing, vidgen-etal-2020-detecting, 10.1145/3487351.3488324, goffredo-etal-2022-counter, 10.3389/frai.2022.932381, lin-etal-2022-multiplex}, and only a few are retrieved from Youtube \citep{mathew2019thou, kennedy2020constructing, priyadharshini-etal-2022-overview} and Reddit \citep{kennedy2020constructing, vidgen-etal-2021-introducing, lee2022elf22, yu-etal-2022-hate}, respectively (though again it is worth noting that at the time of writing the Twitter API was starting to become a lot less accessible). 

In the second category, counterspeech is written by crowd workers \citep{qian-etal-2019-benchmark} or operators expert in counterspeech writing \citep{chung-etal-2019-conan, chung-etal-2021-towards}. While such an approach is expected to offer relatively controlled and tailored responses, writing counterspeech from scratch is time-consuming and requires human effort. To address this issue, advanced generative language models are adopted to automatically produce counterspeech \citep{tekiroglu-etal-2020-generating, fanton-etal-2021-human, bonaldi2022dialoconan}, as we will discuss further below.

Regarding granularity of taxonomies, most existing datasets provide binary annotation (counterspeech/non-counterspeech) \citep{garland-etal-2020-countering, vidgen-etal-2020-detecting, 10.1145/3487351.3488324, vidgen-etal-2021-introducing}, while three datasets feature annotations of the types of counterspeech \citep{mathew2018analyzing, mathew2019thou, chung-etal-2019-conan}. In terms of hate incidents, datasets are available for several hate phenomena such as islamophobia \citep{chung-etal-2019-conan} and East Asian prejudice during COVID-19 pandemic \citep{vidgen-etal-2020-detecting, 10.1145/3487351.3488324}. The aforementioned datasets are mostly collected and analyzed at the level of individual text, not at discourse or conversations (e.g., multi-turn dialogues \citep{bonaldi2022dialoconan}). Most of the datasets are in English, while only a few target multilinguality, including Italian \citep{chung-etal-2019-conan, goffredo-etal-2022-counter}, French \citep{chung-etal-2019-conan}, German \citep{garland-etal-2020-countering}, and Tamil \citep{priyadharshini-etal-2022-overview}.

\subsection{Approaches to Counterspeech Detection}

Previous work on counterspeech detection has focused on binary classification (i.e. whether a text is counterspeech or not) \citep{vidgen-etal-2020-detecting, garland2022impact, 10.1145/3487351.3488324} or identifying the types of counterspeech as a multi-label task \citep{mathew2018analyzing, garland-etal-2020-countering, chung-etal-2021-multilingual, goffredo-etal-2022-counter}. Automated classifiers are developed to analyse large-scale social interactions of abuse and counterspeech addressing topics such as political discourse \citep{garland2022impact} and multi-hate targets \citep{mathew2018analyzing}. Moving beyond monolingual study, \citet{chung-etal-2021-multilingual} evaluate the performance of pre-trained language models for categorising counterspeech strategy for English, Italian and French in monolingual, multilingual and cross-lingual scenarios.

\subsection{Approaches to Counterspeech Generation}
Various methodologies have been put forward for the automation of counterspeech generation \citep{qian-etal-2019-benchmark}, addressing various aspects including the efficacy of a hate countering platform \citep{CHUNG2021100150}, informativeness \citep{chung-etal-2021-towards}, multilinguality \citep{chung2020italian}, politeness \citep{Saha2022}, and grammaticality and diversity \citep{zhu-bhat-2021-generate}. These methods are generally centred on transformer-based large language models (e.g., GPT-2 \citep{radford2019language}). By testing various decoding mechanisms using multiple language models, \citet{tekiroglu-etal-2022-using} find that autoregressive models combined with stochastic decoding yield the optimal counterspeech generation. In addition to tackling hate speech, there are studies investigating automatic counterspeech generation to respond to trolls \citep{lee2022elf22} and microagressions \citep{ashida-komachi-2022-towards}.

\paragraph{Evaluation of counterspeech generation}
Assessing counter speech generation is complex and challenging due to the lack of clear evaluation criteria and robust evaluation techniques. 

Previous work evaluates the performance of counterspeech systems via two aspects: automatic metrics and human evaluation. 
Automatic metrics, generally, evaluate the generation quality based on criteria such as linguistic surface \citep{papineni2002bleu, lin-2004-rouge}, novelty \citep{wang2018sentigan}, and repetitiveness \citep{bertoldi2013cache, cettolo2014repetition}. Despite being scalable, these metrics are uninterpretable and can only infer model performance according to references provided (e.g., dependent heavily on exact word usage and word order) and gathering an exhaustive list of all appropriate counterspeech is not feasible. For this reason, such metrics cannot properly capture model performance, particularly for open-ended tasks \citep{liu-etal-2016-evaluate, novikova-etal-2017-need} including counterspeech generation. 
As a result, human evaluation is heavily employed based on aspects such as suitableness, grammatical accuracy and relevance \citep{chung-etal-2021-towards, zhu-bhat-2021-generate}. Despite being trusted and high-performing, human evaluation has inherent limitations such as being costly, difficult (e.g., evaluator biases and question formatting), and time-consuming (both in terms of evaluation and moderator training), and can be inconsistent and inflict psychological harm on the moderators. 
The effectiveness of counterspeech generations should be also carefully investigated `in-the-wild' to understand its social media impact, reach of content, and the dynamics of hateful content and counterspeech (see Section \ref{sec:cs_impact}). To our knowledge, no work has examined this line of research yet.

\paragraph{Potentials and limits of existing generative models }

We believe that in some circumstances counterspeech may be a more appropriate tool than content moderation in fighting hate speech as it can depolarise discourse and show support to victims. 
However, automatic counterspeech generation is a relatively new research area. Recent progress in natural language processing has made large language models a popular vehicle for generating fluent counterspeech. However, counterspeech generation currently faces several challenges that may constrain the development of efficient models and hinder the deployment of hate intervention tools. Similar to the use of machine translation and email writing tools, we advocate that counterspeech generation tools should be deployed as suggestion tools to assist in hate countering activity \citep{chung-etal-2021-towards, CHUNG2021100150}.

\vspace{0.2cm}

\noindent\textbf{$\bullet$ Faithfulness/Factuality in generation} Language models are repeatedly reported to produce plausible and convincing but not necessarily faithful/factual statements \citep{solaiman2019release, NEURIPS2019_3e9f0fc9, chung-etal-2021-towards}. We refer to faithfulness as being consistent and truthful in adherence to the given source (i.e. model inputs) \citep{10.1145/3571730}. Many attempts have been made to mitigate this issue \citep{10.1145/3571730}, including correcting unfaithful data \citep{nie-etal-2019-simple}, augmenting inputs with additional knowledge sources \citep{chung-etal-2021-towards}, and measuring faithfulness of generated outputs \citep{dusek-kasner-2020-evaluating, zhou-etal-2021-detecting}. We encourage reporting the faithfulness/factuality of models.

\vspace{0.2cm}

\noindent\textbf{$\bullet$ Toxic degeneration} Language models can also induce unintendedly biased and/or toxic content, regardless of whether explicit prompts are used \citep{dinan-etal-2022-safetykit}. In the use case of counterspeech generation, this can result in harm to victims and bystanders as well as risking provoking perpetrators into further abusive behaviour. This issue has been mitigated by two approaches: data and modelling. The data approach aims at creating proper datasets for fairness by removing undesired and biased content \citep{blodgett-etal-2020-language, JMLR:v21:20-074}. The modelling approach focuses on controllable generation techniques that, for instance, employ humans for post-editing \citep{tekiroglu-etal-2020-generating} and detoxification techniques \citep{gehman-etal-2020-realtoxicityprompts}. 

\vspace{0.2cm}

\noindent\textbf{$\bullet$ Generalisation vs. Specialisation}
With the rise of online hate, models that can generalize across domains would be helpful for producing counterspeech involving new topics and events. Generalisable methods can also ameliorate the time and manual effort required for collecting and annotating data. However, as discussed in Section \ref{sec:cs_impact}, counterspeech is multifaceted and contextualised. There may not be a one-size-fits-all solution. For instance, abuse against women can often be expressed in a more subtle form as microaggressions. 
It may, therefore, be difficult to implement an easy yet effective counterspeech strategy in one model. Moreover, model generalisability is challenging \citep{FORTUNA2021102524, yin2021towards}, and can have potential limitations \citep{conneau-etal-2020-unsupervised, berend-2022-combating}. Finding the right trade-off between generalisation and specialisation is key. 

\section{Future Perspectives}
Of the many promising abuse intervention experiments that we review, results are not always consistent, demonstrating weak claims or limited success (applicable only to certain settings). Possible reasons include short-term experiments, small sample sizes and non-standardised experimental designs. 
To improve this, effective interventions should come with the characteristics of scalability, durability, reliability, and specificity. In this section, we highlight key distinctions and overlaps across areas that have and have not been explored in social sciences and computer science, discuss ethical issues related to evaluating counterspeech in real-life settings and automating the task of counterspeech generation, and identify best practices for future research.

\paragraph{Distinctions and overlaps across areas} 
By recognizing the commonalities and differences between social sciences and computer science, we pinpoint the unique contributions of each discipline and encourage interdisciplinary collaborations to address complex societal challenges and better understand human behaviour with the help of computational systems.

\vspace{0.2cm}

\textbf{$\bullet$ Terminological clarity.} Throughout the counterspeech literature, terminology is used inconsistently. Terms such as counterspeech and counter-narratives are often used interchangeably or used to refer to similar concepts. In social science, counterspeech is used to refer to content that disagrees with abusive discourses and counter-narratives often entail criticism of an ideology with logical reasoning. As a result, counter-narrative stimuli designed in social experiments are generally long form \citep{10.3389/fpsyg.2020.01059}. In computer science on the other hand, the distinctions between counterspeech and counter-narratives have been vague, and training data is generally short form (while this may be bound by character limit on social media platforms). For instance, short and generic responses such as `\textit{How can you say that about a faith of 1.6 billion people?}' can be commonly found in counter-narrative datasets \citep{chung-etal-2019-conan}.

\vspace{0.2cm}
\noindent\textbf{$\bullet$ The focus of evaluation.} Social scientists and counterspeech practitioners generally attempt to understand and assess the impact of counterspeech on reducing harms (e.g., which strategies are effective and public perception towards counterspeech), whereas computer scientists focus more on technical exploration of automated systems and testing their performance in producing counterspeech (e.g., comparing system outputs with a pre-established ground truth or supposedly ideal output). One commonality between the social science and computer science studies is that most findings are drawn from controlled and small-scale studies. Applying interventions to real-world scenarios is a critical next step.

\vspace{0.2cm}
\noindent\textbf{$\bullet$ Datasets.} Dataset creation is an important component in computer science for developing machine learning models for generating counterspeech, while such contributions are less commonly considered in social sciences which rely on experiments using hand-crafted stimuli and one-time analyses of their effectiveness. 

\vspace{0.2cm}
\noindent\textbf{$\bullet$ Scope of research.} We observe that, while computer scientists have focused on responses to abusive language and hate speech, social science studies address a wider range of phenomena, in particular radicalisation and terrorist extremism. 
It can be difficult to measure the effectiveness of counterspeech in challenging these over the short term, leading to some of the differences in evaluation metrics across disciplines.

\vspace{0.2cm}
\noindent\textbf{$\bullet$ Lack of standardised methodologies.} 
A variety of methodologies have been adopted in the literature, making comparisons across studies difficult. Without standardised evaluations, it is difficult to situate the results and draw robust findings.

\paragraph{Ethical Issues, Risks and Challenges of Conducting Counterspeech Studies}
Effective evaluation of counterspeech not only identifies users who may need help, but also safeguards human rights and reinforces a stronger sense of responsibility in the community. 
This discussion is based on the authors' opinion and not stemming from the review.

\vspace{0.2cm}

\noindent\textbf{$\bullet$ Evaluating counterspeech in real-life settings} Conducting the evaluation of counterspeech in real-world scenarios appears to provide a proactive and quick overview of its performance on hate mitigation. However, from an ethical perspective, the debate surrounding it is ongoing and reaching an agreement can be difficult. 
For instance, one side argues about the morality of exposing participants to harm, while another points to the importance of internet safety. Exercising counterspeech can offer mitigation of online abuse in good faith and may be exempt from liability based on several legal groundings. As an example, \textbf{Good Samaritan laws} provide indemnity to people who assist others in danger \citep{smits2000good}. In 2017 the EU Commission released a communication on tackling illegal content online, stating that `\textit{This Communication ... aims to provide clarifications to platforms on their liability when they take proactive steps to detect, remove or disable access to illegal content (the so-called ``Good Samaritan'' actions)}' \citep{european2017communication}. Section 230(c)(2) of Title 47 of the United States Code extents this protection to the good faith removal or moderation of third-party material they deem ``obscene, lewd, lascivious, filthy, excessively violent, harassing, or otherwise objectionable, whether or not such material is constitutionally protected.'' and stresses liability towards online hate speech. It protects online computer services from liability for moderating third-party materials that are harmful \citep{ardia2009free, goldman2018overview}. The aim of these safeguards is to ensure that individuals are not hesitant to help others in distress due to the fear of facing legal consequences in case of unintentionally making errors in their efforts to provide support.

Responsible open-source research can facilitate reproducibility and transparency of science. Recently, reproducible research has been deemed critical in both social sciences \citep{10.2307/44282622, doi:10.1177/17456916211041116} and computer science, and low replication success is found despite using materials provided in the original papers \citep{belz-etal-2023-missing,doi:10.1126/science.aac4716}. To tackle this issue, a few initiatives for transparent research have been proposed, advocating researchers to state succinctly in papers how experiments are conducted (e.g., stimuli, mechanisms for data selection) and evaluated, including A 21 Word Solution \citep{simmons201221} and Open Science Framework.\footnote{\url{https://osf.io/}} Furthermore, practising data sharing encourages researchers to be responsible for fair and transparent experimental designs, and to avoid subtle selection biases that might affect substantive research questions under investigation \citep{dennis2019privacy}. At the same time, when handling sensitive or personal information, data sharing should adhere to research ethics and privacy standards \citep{dennis2019privacy, de2022open}. For instance, in the case of hate speech, using synthetic examples or de-identification techniques is considered a good general practice for ensuring the safety of individuals \citep{kirk-etal-2022-handling}.

\vspace{0.2cm}

\noindent\textbf{$\bullet$ Automating counterspeech generation} There are several ethical challenges related to automating the task of counterspeech generation.
First of all, there is the danger of dual-use:
the same methodology could also be used to silence other voices.

Furthermore, effective and ethical counterspeech relies on the accuracy and robustness  of detecting online hate speech: an innocent speaker may be publicly targeted and shamed if an utterance is falsely classified as hate speech -- either directly or indirectly as in end-to-end response generation.
For example, Google's Jigsaw API \citep{perspectiveAPI}, a widely used tool for detecting toxic language, makes predictions that  are aligned with racist beliefs and biases---for example it is less likely to rate anti-Black language as toxic, but more likely to mark African American English as toxic \citep{sap-etal-2022-annotators}.
It is thus important to make sure that the underlying tool is not biased and well-calibrated to the likelihood that an utterance was indeed intended as hate speech. For example, the `tone' of counterspeech could be used to reflect the model's confidence.

A related question is free speech: what counts as  acceptable online behaviour, what sort of speech is deemed inappropriate, in which contexts, and should be targeted by counterspeech? A promising direction for answering this complex question is participatory design to empower the voices of those who are targeted \citep{birhane-etal-2022-power}.

In sum, there is a trade-off between risks and benefits of counterspeech generation. Following the `Good Samaritan' law:  automating counterspeech provides timely help to victims in an emergency which is protected against prosecution (even if it goes wrong). Similar legislation is adopted by other countries, including the European Union, Australia and the UK. Under this interpretation, well-intentional counterspeech (by humans and machines) is better than doing nothing at all.

\paragraph{Best practices}

We provide best practices for developing successful intervention tools.

\begin{enumerate}
  \item Bear in mind practical use cases and scenarios of hate-countering tools. A single intervention strategy is unlikely to diminish online harm. To design successful counterspeech tools, it is important to consider the purposes of counter messages (e.g., support victims and debunk stereotypes), the speakers (e.g., practitioners, authorities and high-profile people), recipients (e.g., ingroup/outgroup, political background and education level), the content (e.g., strategy, style, and tones), intensity (e.g., one message per week/month), and the communication medium (e.g., videos, text, and platforms). 
  \item Look beyond automated metrics and consider deployment settings for evaluating the performance of generation systems. Generation systems are generally evaluated on test sets in a controlled environment using accuracy-based metrics (e.g., ROUGE and BLEU) that cannot address social implications of a system. Drawn from social science studies, metrics assessing social impact (e.g., user engagement), behavioural change (e.g., measure abuse reduction in online discourse) and attitude change (e.g., through self-description questionnaires) can be considered. A good intervention system is expected to pertain long-lasting effects.
  \item Be clear about the methodology employed in experiments, open-source experimental materials (e.g., stimuli, questionnaires and codebook), and describe the desirable criteria for evaluating counterspeech intervention. As standardised procedures are not yet established for the assessment of counterspeech interventions, examining the impact of interventions becomes difficult. A meaningful description of experimental design would therefore enhance reproducible research and help capture the limitation of existing research.
  \item Establish interdisciplinary collaboration across areas such as counter-terrorism, political science, psychology and computer science. AI researchers can help guide policymakers and practitioners to, for instance, identify long-term interventions by performing large-scale data analysis using standardized procedures on representative and longitudinal samples. With expertise in theories of human behaviour change and experimental design, social science researchers can conduct qualitative evaluations of AI intervention tools in real-life scenarios to understand their social impact. 
\end{enumerate}

 \section{Conclusion}
 Online hate speech is a pressing global issue, prompting scientists and practitioners to examine potential solutions. Counterspeech, content that directly rebuts hateful content, is one promising avenue. While AI researchers are already beginning to explore opportunities to automate the generation of counterspeech for the mitigation of hate at scale, research from the social sciences points to many nuances that need to be considered regarding the impact of counterspeech before this intervention is deployed. Taking an interdisciplinary approach, we have attempted to synthesize the growing body of work in the field. Through our analysis of extant work, we suggest that findings regarding the efficacy of counterspeech are highly dependent on several factors, including methodological ones such as study design and outcome measures, and features of counterspeech such as the speaker, target of hate, and strategy employed. While some work finds counterspeech to be effective in lowering further hate generation from the perpetrator and raising feelings of empowerment in bystanders and targets, others find that counterspeech can backfire and encourage more hate. To understand the advantages and disadvantages of counterspeech more deeply, we suggest that empirical research should focus on testing counterspeech interventions in real-world settings which are scalable, durable, reliable, and specific. Researchers should agree on key outcome variables of interest in order to understand the optimal social conditions for producing counterspeech at scale by automating its generation. We hope that this review helps make sense of the variety of types of counterspeech that have been studied to date and prompts future collaborations between social and computer scientists working to ameliorate the negative effects of online hate. 

\section*{Acknowledgements} We thank Bertie Vidgen for the valuable feedback on the initial structure of this manuscript and Hannah Rose Kirk for her help with the collection of target literature.

\bibliographystyle{apalike}
\bibliography{custom}

\begin{thebibliography}{}
\bibitem[Commission, 2017]{european2017communication}
Commission, E. (2017).
\newblock Communication from the commission to the european parliament, the
  council, the european economic and social committee and the committee of the
  regions.
\end{thebibliography}

\end{document}